\documentclass[runningheads]{llncs}
\usepackage{graphicx}
\usepackage{url}
\usepackage{array}
\usepackage{verbatim}
\usepackage{listings}
\usepackage{relsize}

\newcolumntype{M}[1]{>{\centering\arraybackslash}m{#1}}
\begin{document}
\title{Towards the Generation of Musical Explanations with GPT-3}
%
%
\author{Stephen James Krol \and
Maria Teresa Llano \and
Jon McCormack}
\authorrunning{S. J. Krol et al.}
%
\institute{Monash University, Melbourne, Victoria \\
\email{stephen.krol@monash.edu}}
\maketitle              
\begin{abstract}
Open AI's language model, GPT-3, has shown great potential for many NLP tasks, with applications in many different domains.  In this work we carry out a first study on GPT-3’s capability to communicate musical decisions through textual explanations when prompted with a textual representation of a piece of music. Enabling a dialogue in human-AI music partnerships is an important step towards more engaging and creative human-AI interactions. Our results show that GPT-3 lacks the necessary intelligence to really understand musical decisions. A major barrier to reach a better performance is the lack of data that includes explanations of the creative process carried out by artists for musical pieces. We believe such a resource would aid the understanding and collaboration with AI music systems.



\end{abstract}

\section{Introduction} 
Human-machine communication in creative interactions is shallow as ultimately the creative processes of both humans and machines are carried out in isolation. On the one hand, machines have a very limited understanding of how their contributions are being received by their users. Conversely, users (and other stakeholders) are left questioning the reasoning behind a system's decisions, confounded by its behaviour and even discouraged from collaborating with it.  Researchers and users of creative systems are increasingly seeking partnerships that not only passively support human creativity but can also enhance it through a more active role in the creative process. This partnership would be one in which systems take the initiative to produce novel creative outputs, explain them and discuss them with their contributors \cite{mccormack:os2020}. Some researchers in the field are pursuing this aim under the lenses of the field of Explainable AI (XAI). For instance, in \cite{zhu:2018}, different levels of explainability to support game designers are identified, while in \cite{xcc:llano}, a framework for bidirectional human-machine communication is proposed. We build upon this work, which explores the role of Explainable AI in creative settings, by undertaking a first study in the generation of explanations of musical decisions made by AI music systems. 

In this work, we leverage the capabilities of transformer-based technologies to generate explanations of musical decisions using GPT-3 \cite{brown:newrips2020}, a state of the art natural language model. GPT-3's ability to generate natural language texts that resemble texts written by humans has made numerous headlines in the media and has attracted the interest of researchers in different areas. As of Spring of 2021, Open AI's blog claimed GPT-3 had been used in over 300 different applications\footnote{\url{https://openai.com/blog/gpt-3-apps/?utm_campaign=}}. In this paper we perform the first study that uses GPT-3 to generate explanations for musical decisions. For this, one of the authors composed a number of musical pieces and provided explanations about their creative process and musical decisions. We used this data to carry out a set of experiments in which we fed GTP-3 with example pieces of the music, in the form of some textual representation, alongside some of the explanations provided by the author. We then tested GPT-3's capability to produce explanations itself. 

Our experiments show that for the context of generating explanations of musical decisions, GPT-3's few-shot learning capabilities are still limited. Even though it is capable of expressing explanations about musical notation, it lacks the necessary intelligence to really understand it, making it unreliable. However, we see potential in fine-tuning GPT-3 or training an open source language model, with a dataset of explanations of musical pieces. In addition to aiding in the understanding and use of AI music systems, we see the role of explanations aiding the creative process itself if used for musical ideation.   


Next we provide an overview of current work on the use of transformer-based language models in XAI and in the context of music. Then we describe a series of experiments that we carried out, which aimed at i) testing GPT-3's knowledge of music theory when providing explanations, and ii) testing its capability to provide explanations about other aspects of the creative process. We end the paper with a discussion of the results and outline possible paths for future work. 
%
%
%

\section{Background} 

Communication is an intrinsic element when collaborating in any creative domain. Musicians, for instance, communicate not only through the music they are playing, but also through visual cues and verbal indications \cite{bishop:2018,bishop:2019}. Providing communication channels that aid collaboration in human-machine interactions in the context of music is an active area of research in the field. We explore current work in this area in the following sections.


\subsection{Communication in human-machine music interactions}
Current research seeks to understand how communication can be enabled in human-machine interactions in the context of music. One focus to this research has been in the development of expressive robotic improvising musicians, with the musical improvisation robot {\em Shimon} being the most notable work  \cite{Hoffman2011,bretan2017towards}. The physical embodiment of {\em Shimon}, through movement and animated appearance, is one of the most important aspects of the research behind this work. Bi-directional communication has also been a subject of research in this area. For instance, McCormack et al \cite{mccormack2019silent} equipped both an AI musician and the human performers with the ability to continuously communicate how confident they felt during an improvised performance. This was achieved by using a visualiser to convey the system's confidence through an emoticon style face, while the human performers communicated their confidence through biometric signals. The work showed that this type of simple, interpretable communication increased the flow within the human-AI collaboration and the quality of the music produced.

Equipping systems with explainable capabilities that aim to reveal aspects of the creative process would also facilitate collaboration in human-machine interactions; however, work on the role of XAI for music, or other creative domains, is scarce. As shown in recent surveys that have attempted to identify the most predominant domains in which XAI is being applied, the domain of Creative AI is either absent \cite{Abdul:2018}, or significantly under represented \cite{anjomshoae:aamas19}, with just a few isolated instances in the domain of game design \cite{zhu:2018,guzdial:2018}. Focused research on the role of explainability for music systems has yet to be widely explored.

\subsection{Transformer-based approaches in music}
The popularity of machine learning approaches in musical contexts has been increasing in recent years. One which has taken the interest of researchers is the use of transformer-based language models for different musical-related tasks. For instance, previous work compared the performance of different transformer-based approaches for composer style classification \cite{tsai:ISMIR20} and instrument classification \cite{ji:ICASSP21} by converting raw images of piano sheet music to text and treating these problems as text classification tasks. In \cite{agrawal:AIR21}, the XLNet language model is used to recognise emotion from lyrics, while in \cite{geerlings:NLP4MUSA20} GPT-2 is used to generate musical sequences in ABC notation. In \cite{d'Eon:NeurIPS20} the authors developed a speech to music composition tool, while in \cite{hsu:2021} a transformer-based model is used for generating music transitions between music clips. Even Open AI has itself produced a model, called MuseNet \footnote{\url{https://openai.com/blog/musenet/}}, which uses transformer-based technology to generate 4-minute musical compositions with different instruments. 

Work using GPT-3 in the musical context is not very vast to the best of our knowledge. This may be due to it not being open-source, which limits researchers in the field to train their own models. However, some musicians and researchers have documented the use of GPT-3 for music. In \cite{santoro:medium}, GPT-3's knowledge about music is tested on different aspects, showing that it is capable of making song recommendations based on time signatures, sound physics, tools and products needed for musical production. However, the creative capabilities of GPT-3 are only tested as far as creating song lyrics in the style of well-known artists. In \cite{gonsalves:medium}, on the other hand, the author developed AI-Tunes, a GPT-3 based tool for creating music. Here GPT-3 is fine-tuned in the task of generating a song when prompted by a song title and band name based on a public dataset, OpenEWLD \footnote{OpenEWLD dataset: \url{https://github.com/00sapo/OpenEWLD}}, which has over 500 songs written in MusicXML format. The songs are converted to the ABC format, which provides a simpler format for training. This work tested a more creative use of GPT-3 for the musical context, with promising results as concluded by the author's evaluation, which was based on identifying tonality features of the generated songs. 

Even though there is a vast application of transformer-based approaches in musical contexts, none of these seek to generate explanations of musical decisions. A reason could be that, while there are many datasets that can be used for the generation of musical outputs, to the best of our knowledge, there are no datasets that have explanations about the creative process and the musical decisions made when composing the pieces.

\subsection{Transformer-based approaches in Explainable AI}
Research that aims at understanding how transformer-based language models behave in order to provide human-understandable interpretations has been focused on analysing the weights of their attention mechanisms \cite{jesse:2019,kovaleva:2019}. Although these explanations provide insights about the decisions behind the model, they fail to provide enough semantic meaning with respect to the creative process. In \cite{bacco:2021}, an attempt is made to provide deeper explanations in the task of sentiment classification. This also involved an analysis of the attention weights; however, instead of focusing the explanations on this feature, they provided a summary of the most relevant sentences (according to the weights) as the explanation. This takes the focus from explaning the model to explaining the task at hand.  

Other approaches are studying the possibility of using external knowledge sources (e.g. WordNet, ConceptNet, etc.) in order to improve the explainability of Natural Language Inference (NLI) models \cite{schuff:2021}. Here we follow a similar approach, carrying out an initial study on the use of language models, particularly GPT-3, as a external source to generate explanations of musical decisions. Although GPT-3 is not used here for generating the music itself, we believe that adapting a music system with an explainable component of this nature can improve the engagement of users. On the one hand, musicians may find that understanding the system will allow them to collaborate easier with it. Additionally, audience members may find it engaging to be able to establish a dialogue with the system where they can unpack aspects of a performance.


\section{Musical Capability of GPT-3}

Although GPT-3 was not specifically trained to understand music theory, it can still communicate useful musical information. However, the extent and reliability of this information is limited. In the following experiments, we investigate GPT-3’s ability to perform musical tasks, such as extracting the key belonging to a sequence of notes and explaining the musical decisions of a fictional song. We also study the use of MusicABC notation with GPT-3 and test the models’ ability to extract musical information from this format. The OpenAI API offers four base models: davinci, curie, babbage and ada. As stated by OpenAI, Davinci is generally the most capable model and is the recommended base model while experimenting. For these reasons, we utilise Davinci in all our models. Additionally, the OpenAI API offers different presets that can be used when interacting with the model. These presets include Q\&A, chat, classification ect. Most experiments were run using the Q\&A format as we felt this was an effective way to limit the randomness of the output.

\subsection{Extracting the key from a sequence of notes}

Determining the key to a song requires a fundamental understanding of music theory and can be an important component when analysing music. This is a task most musicians can do when given enough information. To test GPT-3’s ability to determine a key from a sequence of notes, we used a very simple form of music notation. Each note was represented by their alphabetic character with the octave of the note being ignored. If the note was sharp or flat a $\sharp$ or \textit{b} token was placed to the right of the note. Timing of the notes was also ignored. This notation is consistent with many music forums, particularly on the subject of key. Below are two prompts that were used in most of the experiments. 
\newline
\newline
Prompt 1: 
{\smaller
\begin{verbatim}
You are a musical assistant that is given a sequence of 
musical notes as input and outputs the key the notes are in.
Q: What key is the following sequence of musical notes in?
C D E F G A B
A: C Major
Q: What key is the following sequence of musical notes in?
G A B C D E F# G
A: G Major
Q: What key is the following sequence of musical notes in?
D E F# G A B C#
A: D major
\end{verbatim}
}

\noindent Prompt 2:
{\smaller
\begin{verbatim}
You are a musical assistant that is given a sequence of 
musical notes as input and outputs the key the notes are in.
Q: What key is the following sequence of musical notes in?
C D E F G A B
A: C Major
Q: What key is the following sequence of musical notes in?
D F# G C D A G
A: G Major
Q: What key is the following sequence of musical notes in?
A C E D G F A
A: A minor
Q: What key is the following sequence of musical notes in?
D F C E G Bb D
A: D minor}
\end{verbatim}
}

As can be seen the prompts provide contextual information and example questions and answers. Table \ref{tab:key-answers} contains the results from the experiment. In these experiments we asked two types of questions. The first were simple questions where the sequence of notes was in the order of the key’s scale. The second were more complex questions where the sequence did not start on the tonic but contained all the relevant notes of the key.

In table \ref{tab:key-answers} we see that while GPT-3 demonstrates some ability to infer the key from a sequence of notes, it cannot do it reliably. In these experiments the model relied heavily on the first note in the sequence to determine the key and consistently failed once the order of notes was shuffled. Furthermore, the design of the initial prompt had a noticeable impact on its performance. For example, if a prompt provided examples of only major keys, GPT-3 would regularly classify a minor sequence as a major sequence.

To further test GPT-3’s knowledge on key, we inverted the problem and asked it to provide a sequence of notes given the key. The results of this can be seen in table \ref{tab:key-generate}. Here GPT-3 is much more reliable and showcases a basic understanding of key. Comparing both experiments we see that while it can recite the notes in a particular key, it cannot reliably infer the key from a sequence of notes. This demonstrates a limitation in GPT-3’s ability to understand written music.

    \begin{table}
        \centering
        \caption{Questions and corresponding answers on key.}
        \label{tab:key-answers}
        \begin{tabular}{|M{2cm}|p{6cm}|c|c|}
        \hline
        \textbf{Prompt Number} & \textbf{Question} & \textbf{Key} & \textbf{GPT3 Answer} \\
        \hline
        1 & 
        Q: What key is the following sequence of musical notes in? A B C\# D E F\# G\#  & A Major & A Major  \\
        \hline
        1 &
        Q: What key is the following sequence of musical notes in? E F\# G\# A B C\# D\#
        & E Major & E Major \\
        \hline
        1 &
        Q: What key is the following sequence of musical notes in? A C\# D G\# E D C\# A & A Major & A Major \\
        \hline
        1 & 
        Q: What key is the following sequence of musical notes in? A B C D E F G 
        & A Minor & A Major \\
        \hline
        1 &
        Q: What key is the following sequence of musical notes in? E B C\# D E F\# G\# A & A Major & E Major \\
        \hline
        2 &
        Q: What key is the following sequence of musical notes in? A E F G D C B A 
        & A Minor & A Major \\
        \hline
        2 &
        Q: What key is the following sequence of musical notes in? D C F\# E A B G
        & G Major & D Minor \\
        \hline
        2 &
        Q: What key is the following sequence of musical notes in?G Bb A C D Eb F G
        & G Minor & G Major \\
        \hline
        \end{tabular}
    \end{table}

    \begin{table}
        \caption{GPT-3 generating notes from key}
        \label{tab:key-generate}
        \centering
        \begin{tabular}{|c|c|c|}
        \hline
        \textbf{Question} & \textbf{Answer} & \textbf{Correct} \\
        \hline    
        Q: What are the notes in E major? & A: E F\# G\# A B C\# D\# E & yes \\
        \hline
        Q: What are the notes in G minor? & A: G A Bb C D E F G & no \\
        \hline
        Q: What are the notes in D minor? & A: D E F G A Bb C D & yes \\
        \hline
        Q: What are the notes in C minor? & A: C D Eb F G Ab Bb C & yes \\
        \hline
        Q: What are the notes in Eb major? & A: Eb F G Ab Bb C D Eb & yes \\
        \hline
        \end{tabular}
    \end{table}

\subsection{Providing explanations for a fictional song}

To further test GPT-3’s understanding of music, we asked questions about musical decisions it had made in a fictional song. We tried three different prompts and designed questions that investigated different musical techniques such as tempo, key and use of dynamics. The first prompt was “You are a musician who is answering questions about their composition”. Questions and answers to this prompt can be seen in table \ref{tab:fictional-1}. These results demonstrate that GPT-3 has a decent understanding of music and can provide logical explanations for musical decisions. It understands the difference in mood between major and minor keys as well as the effect tempo can have on the music. It also understands how different dynamics effect the music and describes how different instruments can be used to express emotion. These results are promising and exhibit GPT-3’s potential to explain musical decisions.

    \begin{table}
        \centering
        \caption{Question and answers related to a non-existent song for the prompt \textit{``You are a musician who is answering questions about their composition''}.}
        \label{tab:fictional-1}
        \begin{tabular}{|p{5cm}|p{7cm}|}
        \hline
        \textbf{Question} & \textbf{Answer} \\
        \hline
         Why did you play with a tempo of 80bpm? & 
         I wanted the song to be sad and depressing. This low tempo allowed me to express this in my music.  \\
        \hline
        Why did you choose a major key for your music? &
        I wanted the song to be happy and uplifting. This major key allowed me to express this in my music. \\
        \hline
        Why did you switch from a major key to a minor key? &
        I wanted to play a song that expressed sadness and this minor key allowed me to express this in my music. \\
        \hline
        Why did you perform this section softly? &
        I wanted to express sadness and this section best suited my mood.\\ 
        \hline
        Why did you use a crescendo here? &
        I wanted to express excitement and this crescendo allowed me to express this in my music. \\
        \hline
        Why did you choose the violin to play this piece of music? &
        I wanted to play a song that expressed sadness and the violin best suited my mood. \\
        \hline
        \end{tabular}
    \end{table}

In the second prompt we aimed to maintain continuity throughout the conversation. Most questions asked were follow-up questions that aimed to push GPT-3. The prompt for these questions was “The following is a conversation with an AI musician. The AI musician has recently finished composing a piece of music and is excited to talk about and explain various aspects about their song”. Questions and answers relating to this prompt can be seen in table \ref{tab:fictional-2}. Here GPT-3 demonstrates more advanced music theory knowledge,is generally consistent with its explanations and can maintain continuity throughout the conversation. However, some of the answers were heavy in music jargon and were difficult to understand.

    \begin{table}[t!]
        \centering
        \caption{Question and answers related to a non-existent song for the prompt \textit{``The  following  is  a  conversation with an AI musician. The AI musician has recently finished composing a piece of music and is excited to talk about and explain various aspects about their song''}.}
        \label{tab:fictional-2}
        \begin{tabular}{|p{5cm}|p{7cm}|}
        \hline
        \textbf{Question} & \textbf{Answer} \\
        \hline
        Tell me about your newest song? & 
        My newest song is a piece called The Matrix Ritual. To compare it to another work of mine, its closest relative is probably Soaring Santorini Sunrise because I use similar instrumentation and have a similar "vibe".   \\
        \hline
        What emotions are you trying to convey in The Matrix Ritual? &
        I am trying to convey that I love learning, creation, learning more. The film the Matrix helps me think about these things, hence the title of the song. \\
        \hline
        How do you convey these emotions? &
        I use modulated strings and bells with what I call a jazz-like swing rhythm. The main drum beats are in syncopated patterns that follow the same rhythm, and there is a constant synth pad in the background that changes melody with each iteration (triplet descending). \\
        \hline
        What key was this song in? &
        It was in E major. \\
        \hline
        Why did you choose this key? &
        Because the melody was in the key of E, and it has 7 flats. \\
        \hline
        What chord progressions did you use in The Matrix Ritual? &
        I used three basic chord progressions: \\
        \hline
        What were these basic chord progressions? &
        The first is a modulation through several modes in the key of C. The modulations are in the sequence I–IV, V–VII, and in modal mixture in the 'crunchy' sections in the middle, I–III–VII. \\
        \hline
        What was the second? &
        a passage through harmonic minor; [it is very] chromatic and uses one of the modes in that scale. \\
        \hline
        and what about the third? &
        [The third is a sectioned off] bit that uses chromatic passages through all of the modes in harmonic minor. \\
        \hline
        \end{tabular}
    \end{table}

The final prompt aimed to limit the complexity of the answers and was “The following is a conversation with an AI musician. The AI musician has recently finished composing a piece of music and is excited to talk about and explain various aspects about their song. The AI musician is describing their song to someone who has little knowledge in music theory. Therefore, the AI musician does not answer with complicated explanations”. Questions and answers to the third prompt can be seen in table \ref{tab:fictional-3}. Here GPT-3 provides explanations that are both detailed and easy to read. Despite not describing a real song, GPT-3 provides explanations for musical techniques that are common in modern music while also maintaining continuity with the overarching theme of the song. Again, showing potential for the generation of explanations of musical decisions.

    \begin{table}[t]
        \centering
        \caption{Question and answers related to a non-existent song for the prompt \textit{``The following is a conversation with an AI musician. The AI musician has recently finished composing a piece of music and is excited to talk about and explain various aspects about their song. The AI musician is describing their song to someone who has little knowledge in music theory. Therefore, the AI musician does not answer with complicated explanations''}.}
        \label{tab:fictional-3}
        \begin{tabular}{|p{5cm}|p{7cm}|}
        \hline
        \textbf{Question} & \textbf{Answer} \\
        \hline
        What is the song about? & 
        Well, it is titled "Blues". It's about love. \\
        \hline
        How did you convey love through your music? &
        I have used some themes related to being in love, but not enough to convey the entire story. \\
        \hline
        What musical elements did you utilize to express love in your song? &
        The use of crescendos represents the build up of excitement when in love. A change in dynamics represents private conversations. The use of counterpoint in my midi-file represents the twist and turns the journey of love takes. \\
        \hline
        What key did you choose for your song? &
        C sharp Minor. \\
        \hline
        Why did you choose C sharp minor? &
        I chose C sharp Minor because the key has a saddening quality to it. I wanted to express longing. The key also fits well with my piece because it contains many sharps in it. \\
        \hline
        What tempo did you choose for the song? &
        I chose 60 bpm as recommended by my song writer program. This tempo is fast enough to convey mood changes, but too slow for the movement to be distracting. \\
        \hline
        \end{tabular}
    \end{table}

\subsection{Extracting musical information using MusicABC notation}
The previous experiments showcased the potential for GPT-3 to provide musical explanations. However, for the model to be used as a method for explaining musical decisions it needs to relate explanations to real written music. The challenge here was to find a textual format that could effectively communicate musical information to the model. The format we chose for this study was MusicABC which was developed by Chris Walshaw \cite{MusicABC:2021} and is a compact text-based music notation system. This format had shown promising results with AI-Tunes \cite{gonsalves:medium} and communicates a lot of musical information in a relatively small number of tokens. We ran a simple experiment to see how well GPT-3 extracts musical information from the MusicABC format. The piece of music used for these experiments was composed by one of the authors of the paper. The questions were simple and only asked for fundamental elements of the song such as the key, timing and use of dynamics. Table \ref{tab:musicABC} contains the results from the experiment. Here GPT-3 performs well and can reliably extract information on the song. Extracting information such as key and timing were simple tasks as these are explicitly stated in MusicABC; however, information on dynamics and octave changes are only present in the written music. This demonstrates the ability of GPT-3 to extract information from the MusicABC format.

    \begin{table}[t]
        \centering
        \caption{Extracting musical information from MusicABC using the prompt \textit{``You are an AI musician who has finished composing a new song. The song is notated in Music ABC notation and can be seen here:} \textbf{\textless song in MusicABC \textgreater} \textit{You are being interviewed by a human who wants to understand this song. Your job is to explain the musical decisions you have made.''}}
        \label{tab:musicABC}
        \begin{tabular}{|p{6cm}|p{4cm}|c|}
        \hline
        \textbf{Question}  &  \textbf{Answer} & \textbf{Correct}\\
        \hline
        What is the key of the song? &
        C Major &  yes\\
        \hline
        What is the time signature of the song? &
        3/4 & yes \\
        \hline
        What instrument is the song meant to played with? &
        Piano & yes \\
        \hline
        Can you describe how you used dynamics in your music? &
        I used dynamics to make the music more interesting. I used a lot of crescendos and decrescendos. & yes \\
        \hline
        What aspects of the song make it a waltz? &
        The time signature is 3/4, and the tempo is 140 beats per minute. & yes \\
        \hline
        Did you change octaves in this song? &
        Yes, I did. I changed octaves in the treble clef. & yes \\
        \hline
        Did you also change octaves in the bass clef? &
        Yes, I did. I changed octaves in the bass clef.& yes \\
        \hline
        \end{tabular}
    \end{table}

\section{Explaining Musical Decisions}

\subsection{Methodology} \label{methodology}
With an understanding of GPT-3’s musical capabilities we began testing its’ ability to generate musical explanations for a precomposed piece of music. To do this, an author of this paper composed a set of short musical pieces. Each song was then annotated with musical explanations and split into multiple sections. Additionally, another experienced musician composed a short piece of music and annotated various sections of it. The explanations were not heavy in music theory and attempted to capture the high-level intentions of the composers (examples of these explanations can be seen in the prompt below). All pieces of music were composed for the piano and then converted to MusicXML using MuseScore3. Once in MusicXML format the songs were converted to MusicABC using the \textit{xml2abc} library \cite{xml2abc:2021}. Each song was then manually split into their various sections by matching each annotation to the relevant measure in the song. 
Prompts were designed to provide important information to the model. This included specifying the notation used, how sections are separated, the general theme of the song and example questions and answers.

\clearpage
\noindent \textbf{Example Prompt: } 
{\smaller
\begin{verbatim}You are an AI musician who has composed a simple piece of music.
The piece of music can be seen below and is in Music ABC notation. 
It is also separated into sections using html tags.
X:1 T:A song %\%score { 1 | 2 } L:1/4 Q:1/4=140 M:3/4 I:linebreak
$ K:C V:1 treble nm="Piano" snm="Pno."
<section1> 
!mp! z3 | z3 | (c3 | B2 d) | c3 | z (B A | c3) | z (d e | d3-) | 
d3 |$ (c3 | B2 d) | c3 | (f2 e) |(e3 | A2 B |!<(! c3 | d2) e!<)!|< 
</section1>
<section2> ... </section2>
<section3> ... </section3>
<section4> ... </section4>
Your inspiration for this song was young love. This was the main
reason you chose to compose a waltz. You will be asked questions
with respect to each individual section.
Q: What is the title and key of your song?
A: My song is titled "A song" and is the key of C major.
Q: Can you describe your musical decisions for section 1?
A: As section 1 is the intro of the song it was important to set
the mood immediately. I decided to play in mezzo piano as I wanted
to set a soft mood that conveyed the innocence of young love.
Q: What was your motivation for section 2? 
A: In section 2 I decided to create some conflict with a sadder melody.
I also used a crescendo to rise from section 1 to section 2
to emphasise the intensity of this conflict. 
\end{verbatim}
}
When evaluating explanations from GPT-3, we looked at how plausible the response was in the context of the music and the composers’ intention. Responses that referred to musical techniques not present in the song or responses that failed to relate to the theme of the song were considered poor explanations. In addition to testing different prompts and questions, we also studied the effect of changing the hyperparameters of the model. The three main parameters we tweaked were temperature, frequency penalty and presence penalty. 


\subsection{Results}
Some output explanations can be seen in table \ref{tab:results}. Detailed transcripts from GPT-3 can be found here\footnote{Study Data: \url{https://github.com/sjkrol/GPT3MusicalExplanationsData}}. Generally, GPT-3 provided poor explanations. At low temperature and frequency values the model would copy phrases used in explanations from different sections. For example, in explanation 4 it states, “I also used a crescendo to rise from section 2 to section 3 to emphasise the intensity of this conflict”. This is directly taken from the composer’s explanation on the transition between section 1 and 2. GPT-3 would also refer to musical techniques not present in the music. For example, at one point it states that it decided “to use staccato notes to create a bouncier melody” even though no staccato was used in that section. GPT-3 also seemed to have difficulty detecting repeating melodies. For example, section 3 was mostly a repeat of section 2 but at a higher octave; however, GPT-3 stated that the melody in section 3 was more complex. These results highlight the unreliability of the model to provide relevant explanations for different sections of written music.

Nonetheless, by tweaking different hyperparameters we found that we could generate better responses from the model. For example, comparing explanations in rows 5 and 6 we see that adjusting temperature, frequency penalty and presence penalty resulted in a more relevant explanation. However, tuning the hyperparameters requires human intervention and does not guarantee an appropriate response. Additionally, the design of the question had an impact on the quality of the explanation. Asking questions about specific sections resulted in more unreliable explanations compared to questions that focus on the overall theme of the song. These results indicate that while GPT-3 shows promise in providing automatic explanations for written music, its’ few-shot learning capability is too inconsistent to generate reliable explanations.
 
    \begin{table}
        \centering
        \caption{Explanations generated by GPT-3.}
        \label{tab:results}
        \begin{tabular}{|p{3cm}|p{5cm}|c|c|c|c|}
        \hline
        \textbf{Question} & \textbf{Answer} & \textbf{Temp} &\textbf{Freq} & \textbf{Pres} & \textbf{Relevant}  \\
        \hline
        Q: Can you describe your musical decisions for section 3? & In section 3 I decided to create a more upbeat melody to contrast with section 2. I also decided to use a decrescendo to create a sense of falling action. &
        0 & 0 & 0 & No \\
        \hline
        Can you describe your musical decisions for section 3? &
        In section 3 I decided to use a more complex melody that was more difficult to play. I also decided to use a diminuendo to create a feeling of sadness. &
        0.5 & 0 & 0 & No \\
        \hline
        Q: Can you describe your musical decisions for section 3? &  I decided to keep the mood of section 3 lighter so it would contrast with section 2. I also decided to use staccato notes to create a bouncier melody. & 0.75 & 0 & 0 & No \\
        \hline
        Why did you repeat the melody of section 2 in section 3 but at a higher octave? & I repeated the melody of section 2 in section 3 to create a sense of familiarity. I also used a crescendo to rise from section 2 to section 3 to emphasise the intensity of this conflict. & 0 & 0 & 0 & No \\
        \hline
        Q: Can you describe your musical decisions for section 4? & I decided to repeat the melody of section 4 in section 3 but in a higher octave and softer dynamic. I did this to return to innocence of section 1 and demonstrate that even this conflict in young love is itself innocent. I also used a diminuendo to transition softly between section 2 and 3. & 0 & 1 & 0 & No \\
        \hline
        Q: Can you describe your musical decisions for section 4? & I decided to use a crescendo in section 4. I did this so that the climax of the song would be heard clearly and also because it was at this point that young love had been resolved, hence why there is no conflict anymore. & 0.4 & 2 & 0.9 & Yes \\
        \hline
        How did you convey young love in your song? & I conveyed young love in my song by using a waltz as the main dance of young lovers. I also used an innocent melody and soft dynamics to convey this feeling. & 0.4 & 2 & 0.9 & Yes \\
        \hline
        How did you convey young love in your song?  &  I decided to use simple melodies with little technical difficulty. I did this because young love is about simplicity and straightforwardness, hence why the key signature of my song was C major (which has no sharps or flats). & 0.72 & 2 & 0.9 & Yes \\
        \hline
        How did you introduce conflict in the song? & I decided to use a dissonance at the beginning of section 2 and gradually bring it back to consonance. I did this so that young love would be contrasted with conflict in such an obvious way. & 0.72 & 2 & 0.9 & Yes \\
        \hline
        \end{tabular}
    \end{table}

\section{Discussion} 
GPT-3 was not trained to generate explanations from written music. Despite this, when providing explanations for a fictional song, it demonstrated an understanding of different musical techniques and how they could be used to convey a common theme. However, its lack of consistency in performance when relating explanations to real pieces of music, makes it unreliable as a mechanism to explain musical decisions. A fundamental reason for this is the lack of data that includes explanations about musical decisions, particularly with respect to aspects such as intention, motivation, struggles, etc., which are intrinsic elements in the creative process. Using extenal knowledge sources has been an effective technique used in the field of XAI; however, our experiments show that this approach does not work in the music context, where a deeper connection between the explanations and the artefacts they are trying to explain is needed. 

Assuming a dataset with musical explanations exists, the model could be fine-tuned to work with and understand written music. This may improve its ability to generate musical explanations. Additionally, the model could be fine-tuned to generate both music and explanations creating a single system that can create and explain its’ creative decisions. Therefore, an immediate future direction for this work is the creation of a resource in which artists record their process and creative decisions together with their musical outputs. There are different benefits of a dataset of this nature: i) artists could train AI systems to understand their creative process at levels current systems can't, ii) systems could be used for educational purposes, where they could critic and advise users by not only focusing on technical details of music theory but also on creative aspects of it, and iii) music practice could be made more accessible, by providing people with different expertise a partner that would provide them with intuitive and easy to interpret insights about their creative practice. 

A promising application of transformer-based technologies is on ideation. As we have illustrated with our experiments, GPT-3 is able to talk about different components of music and it is able to provide recommendations at different levels. For instance, using a slightly modified version of the prompt in section \ref{methodology}, we asked GPT-3 what would be a good title for the song. We got the following answers when prompted to reply to the same question different times:

<<<<<<< HEAD
As the kind of data that is available in music datasets is not suitable for explainability, a direction for this work is the creation of a resource in which artists record their process and creative decisions together with their musical outputs. There are different benefits of a dataset of this nature: i) artists could train AI systems to understand their creative process at levels current systems can't, ii) systems could be used for education purposes, where they could critic and advise users by not only focusing on technical details of music theory but also on creative aspects of it, and iii) music practice could be made more accessible, by providing people with different skills a partner they can feel comfortable with and that would also provide them with intuitive and easy to interpret insights about their creative practice.

Another application could be related to ideation .... [expand on this]
=======
\begin{enumerate}
    \item A good title for the song would be "A song with an intro, a conflict, and a resolution." 
    \item I have a few suggestions. I think that "Young Love" would be a good title for the song. I also think that "Conflict" could be a good title for this song. 
    \item The first section has a very bright and peaceful mood. The title should reflect this feeling.
    \item A young lovers waltz.\end{enumerate}

Even though the first two suggestions are very simple, and the last suggestion appears to be the name of an existing song\footnote{\url{https://www.youtube.com/watch?v=vqNOTcBfjUM}} -- which raises ethical concerns -- the third suggestion highlights relevant features of the song that a composer could use to produce a title, or to keep pushing GPT-3 to produce one. We also asked GPT-3 for specific ideas on the piece of music; for instance through the following prompt: \textit{``Playing in mezzo piano in section 1 sets a soft mood that conveys the innocence of young love. Could you suggest what would convey conflict?''}, for which GPT-3 replied: \textit{``Playing in mezzo forte in section 1 would convey conflict.''}. Again, showing musical knowledge that could be used as part of an AI musical collaborator.

It is important to also highlight the significance of prompt design, which has become an important consideration when using transformer-based technology. Designing effective prompts can be a difficult task, particularly when domain knowledge is needed. In our experiments we found small changes in a prompt can have a noticeable impact on the results. Being able to identify elements that make good prompts is an important direction for research that aims to create an AI musical collaborator using transformer-based technology.
>>>>>>> 0c8fa8abba99acd3d4a58653b135228cdbe8c0c1

Finally, we want to point out that although we focused on GPT-3, as our aim was to test the capabilities of the state of the art language model, other open source language models should be explored; particularly if a dataset as the one described before can be created and used for training. 


\section{Conclusion}
In this paper we investigated GPT-3’s capability to provide explanations for written music. In early experiments the model demonstrated some knowledge in music theory and could generate a sequence of notes for different keys.  When asked to provide explanations for a fictional song, the model generated explanations that were logical and consistent. The explanations referred to common techniques in music such as counter-point and showcased GPT-3’s potential to generate reasonable explanations for music. The model could also extract musical information from a song written in MusicABC notation. 

To test GPT-3’s explanations on real music, we composed a set of songs for the piano and annotated various sections of the music with explanations of our creative decisions. We designed various prompts to prime GPT-3 with and asked it to explain musical decisions for different sections of the song. The results showed that GPT-3 could not reliably generate reasonable musical explanations for the music. Most explanations either referred to musical elements not present in the music or copied explanations from previous sections. As GPT-3 was not specifically designed for this task, a potential solution could involve fine-tuning the model on musical explanations. However, to the best of our knowledge, a musical dataset that includes musical explanations from the composer does not exist. We hypothesize that creating this dataset could assist XAI researchers in creating creative musical systems that can explain their decisions. Future work will involve curating this dataset and training various models to generate music and explain their musical decisions.
%
%
%

\subsubsection{Acknowledgements} The work presented here was funded by an Early Career Researcher Seed grant awarded by the Faculty of IT at Monash
University.

\begin{small}
\bibliographystyle{splncs04}
\bibliography{ref}
\end{small}

\end{document}